\documentclass[conference]{IEEEtran}
\usepackage{cite}
\usepackage{amsmath,amssymb,amsfonts}
\usepackage{graphicx}
\usepackage{textcomp}
\usepackage{xcolor}
\def\BibTeX{{\rm B\kern-.05em{\sc i\kern-.025em b}\kern-.08em
    T\kern-.1667em\lower.7ex\hbox{E}\kern-.125emX}}

\usepackage{hyperref}
\usepackage{url}

\usepackage{algorithm}
\usepackage[noend]{algorithmic}

\usepackage{amsmath}
\usepackage{makecell}

\usepackage{caption}
\usepackage{subcaption}

\title{Deep Reinforcement Learning with Adjustments }

\author{Hamed Khorasgani,  Haiyan Wang, Chetan Gupta, and Susumu Serita \\
\\ Hitachi Industrial AI Lab, Santa Clara, USA
}

\begin{document}

\maketitle

\begin{abstract}
Deep reinforcement learning (RL)  algorithms can learn complex policies  to optimize agent operation over time.  RL algorithms have shown promising results in solving complicated problems in recent years. 
However, 
their application on real-world physical systems remains limited.  Despite  the advancements in RL algorithms,  the industries often prefer traditional control strategies.  Traditional methods are simple, computationally efficient and easy to adjust. In this paper, we first propose a new Q-learning algorithm for continuous action space, which can bridge the control and RL algorithms and bring us the best of both worlds. Our method can learn complex policies to achieve long-term goals and at the same time it can be 
 easily  adjusted to address short-term requirements without retraining. 
Next, we present an approximation of our algorithm which can be applied to address short-term requirements of  any pre-trained RL algorithm.  
  The case studies demonstrate that  both our proposed method as well as its practical approximation    can achieve short-term and long-term goals without complex reward functions. 
\end{abstract}

\section{Introduction}

Optimal control methodologies  use  system dynamic equations  to design actions that minimize  desired cost functions. A cost function can be designed to track a trajectory,  reach  a goal, or avoid   obstacles. It is also possible to design a cost function to achieve a combination of goals. Model Predictive Control  (MPC) is a  common optimal control technique and has been  applied to many industrial applications such as pressure control and temperature control in chemical processes \cite{garcia1989model}. 
The traditional control solutions are not adequate to address the challenges raised with the  evolution of industrial systems. 
Recently, deep reinforcement learning (RL) has shown promising results in solving complex problems. For example, it has generated superhuman performance in chess and shogi \cite{silver2017mastering}. The following advantages make deep RL a strong candidate to overcome  traditional control limitations. First, deep RL has an advantage in solving   complex problems, especially when the consequences of an action are not  immediately obvious. Moreover, it can learn an optimal solution without requiring detailed knowledge of the systems or their engineering designs. 
Finally, deep RL is not limited to time-series sensors and can use new sensors such as vision for a better control. 

However, deep RL has not been applied to address industrial problems in a meaningful way. There are several key issues that limit the application of deep RL to real-world problems \cite{dulac2019challenges}. 
Deep RL algorithms typically require many samples during  training (sample complexity). Sample complexity leads to high computational costs. A high computational cost can be justified for industries as a one-time charge. However,   
  oftentimes small changes in the system goal,  such as  changing the  desired temperature in a chemical reactor, or a new constraint  such as  a maximum allowable temperature in the reactor,   require retraining the model. Moreover, 
  industrial systems often have several short-term and long-term objectives. 
 Designing a reward function that can capture these short-term and long-term goals concurrently can be  challenging or even infeasible.


A class of short-term objectives   related to safe exploration   during RL training   have  been studied   recently. 
Gu et al.  \cite{gu2017deep} presented an application  of deep RL  for  robotic manipulation control. To ensure 
 safe exploration, 
 they set a  sphere boundary for the end-effector position and when the boundaries  were about to be violated, they used  correction velocity to force the end-effector position back to  the center of the sphere. 
Dalal et al. \cite{dalal2018safe}  formulated the safe exploration as an optimization problem. They proposed   to add a safety
layer that modifies the action at each time step.  Toward this end, they learn the  constraint function using a linear model  and use this model to find the minimal
change to the action such that the safety constraints are met at each time step. To the best of our knowledge,  there is no study addressing short-term objectives during  application. 

In this paper, we present a Locally Linear Q-Learning (LLQL) algorithm for continuous action space. The LLQL includes a short-term prediction model, a long-term prediction model, and a controller.  The short-term prediction model represents a locally linear model of the dynamic system, while the long-term prediction model represents the value function + a locally linear advantage function. 
The controller uses the short-term prediction model and the long-term prediction model to generate actions that achieve short-term and long-term goals simultaneously. It adopts a policy that maximizes Q-value while achieving short-term goals. To make the adoption of our algorithm easier, we propose an approximation version of LLQL which can be applied to any pre-trained RL algorithm. 
Our solution has the following advantages:
 \begin{itemize}
 \item  It does not require designing  sensitive reward functions for achieving  short-term and long-term goals concurrently. 
 \item It shows better performance in achieving short-term and long-term goals compared to the  traditional reward modification methods. 
  \item  It is possible to   modify the short-term goals  without time-consuming retraining.
    \item  It can be applied to any pre-trained RL model.
 \end{itemize}

The rest of this paper is organized as follows. Section \ref{sec:Background and Definitions} 
represents the background in dynamic systems and RL. 
Section \ref{sec:Locally Linear Q-Learning} represents the LLQL algorithm.  Section \ref{sec:Short-term and long-term goals} presents our methodology to achieve short-term and long-term goals using LLQL. Section \ref{Approximation} represents LLQL algorithm's approximation for pre-trained RL algorithms. 
Section \ref{sec:Experimental Results} presents our experimental results. 
 Section \ref{sec:Conclusions} presents the conclusions of the paper. Section \ref{sec:Related Work} discusses the related work. 

\section{Related Work}
\label{sec:Related Work}
 Our work can be categorized as a new model-based RL approach. Model-based RL algorithms use the environment model which represents the state transition function to plan ahead and select actions that lead to higher rewards. 
Several model-based algorithms 
assume the environment model is known.   Among them, 
\textit{AlphaZero} \cite{silver2017mastering} is one of the most famous.  
AlphaZero uses the game's rules (Chess, Shogi and Go)  as the environment model to generate a series of self-play simulated games. During the  simulations, the actions for both players are selected using a Monte-Carlo Tree Search (MCTS) algorithm. The MCTS performs as a planning algorithm by  generating  candidate actions which are superior to the current policy. 
The neural network parameters are updated at the end of each game to minimize the game prediction error (loss, draw or win) and maximize the similarity of policy vector to the planning algorithm. AlphaZero is limited to the discrete  action space problems. 
The environment model   is typically unknown in real-world applications.  Therefore, many model-based RL algorithms learn the state transition model from the data.  \textit{NAF}  \cite{gu2016continuous} learns a linear model for state transition at each operating point and uses this model to generate additional samples through imagination rollout. \textit{World Models} \cite{ha2018world} uses a Variational Auto Encoder (VAE) to map a state variable, $x \in X$ to a lower dimensional variable $z$ in a  latent space $Z$. It then uses a recurrent neural network (RNN)  to learn the state transition model in the latent space. Finally, it applies a simple linear controller to $z$ and the hidden state in the RNN, $h$, to control the system.

Imagination-Augmented Agents (\textit{I2As}) \cite{racaniere2017imagination} introduces two paths: 1) model-free path and 2) imagination path. The imagination path learns a transition model  and uses this model to generate imagination rollouts. These rollouts are aggregated with the samples in the model-free path. To generate actions in the imagination path, I2As uses the model-free path policy. Therefore, the rollouts in the imagination path improve as the I2As policy improves. Using the imagination rollouts, I2As  converge faster  than a model-free network with the same number of parameters. 
 \cite{nagabandi2018neural} showed that a two-step control policy based on 1) learning the dynamic model and 2) applying MPC to the learned model is significantly more sample efficient  than model-free RL. However, this approach cannot achieve high rewards. To achieve higher rewards and preserve sample efficiency, they proposed a hybrid model-based and model-free (\textit{MBMF}) algorithm which runs the model-based approach to achieve the initial result in a sample efficient way, it then trains a model-free   policy to mimic the learned model-based controller, and  uses the resulting imitation policy as   the initialization for  the final model-free RL algorithm.

\cite{feinberg2018model} proposed Model-based
Value Expansion (\textit{MVE}) algorithm, which
limits the uncertainty in the model by only allowing imagination up to a fixed number of steps, H. MVE 
uses the learned system dynamic model to generate simulation data  up
 to H steps into the future, and applies these sample points to estimate Q-function.   Instead of saving 
simulated samples in an imagination buffer, MVE retrains the dynamic model  and generates a new imagination rollout at each step.  \cite{buckman2018sample} expanded MVE algorithm by proposing Stochastic Ensemble Value Expansion (\textit{STEVE}), to generate a  solution more robust to model uncertainty.
 \cite{dalal2018safe}  proposed safe exploration by modeling constraints using  a linear model  and applied Lagrangian optimization to modify the action in order to guarantee safety. In this work, we also used Lagrangian optimization for short-term constraints. However, our approach is different in two ways: 1) our method does not modify the RL action to achieve the goals. Instead, it derives an action by considering both long-term goals and short-term constraints. This is possible because our algorithm  
uses a locally linear  model to represent the advantage function. 2) Unlike safe exploration, the focus of this paper is in  handling   new constraints in the application phase  without retraining the model.

\section{Background and Definitions}
\label{sec:Background and Definitions}
In this section, we review the backgrounds in dynamic systems and reinforcement learning.

\subsection{Dynamic  Systems}
A continuous-time dynamic system can be represented as:
 \begin{small} 
\begin{equation}
\begin{aligned}[c]
 \frac{dx (t)}{dt} &=  f(x(t), u(t), t; p), 
\end{aligned}
\label{continuous-time dynamical system}   
\end{equation}
 \end{small} 
where given the system parameters, $p$, $f$ 
maps the  state variables,  $x \in X$,  and  actions, $u \in U$,  to the state derivative, $\frac{dx}{dt}$ at time $t$. In state space control, the goal is to design a control policy, $\pi_{control} (u(t)|x(t))$, that generates proper actions so as the state variables   follow the given desired trajectory,  $x_d(t)$. 
It is challenging to  design a control policy for a nonlinear complex system represented in equation (\ref{continuous-time dynamical system}). 

The control problem becomes much easier to address when this system is linear with respect to the input \cite{chen2003optimal}.
We can present these systems as: 
\begin{equation}\frac{dx (t)}{dt} = f(x(t)) + g(x(t))u(t). \label{linear control}\end{equation}

Since measurements are typically  sampled in discrete times, we derive
a discrete time version of  linear system (\ref{linear control}). Using
a first-order approximation as:
 \begin{small} 
\begin{equation} x_{k+1}= x_k +  \Delta (f(x_k) + g(x_k)u_k), \label{dis linear control} \end{equation}  \end{small} 
where $x_k$ represents variable $x$   at sample point $k$. $\Delta$ is the sampling rate. In this paper, we assume $\Delta$ is constant. 

\subsection{Reinforcement Learning}
The goal of RL is to learn a policy, $\pi_{RL} (u_k|x_k)$,   that generates a set of actions, $u \in U$, that maximize the expected sum of rewards, $R_k=\sum_{i=k} ^{T} \gamma^{i-k} r(x_i, u_i)$, 
where $\gamma <1$ is the discount factor, $r$ is the reward function and  $T$ represents the end time and can be set to $T= \infty$. The goal is to learn $\pi_{RL}$ for environment, $E_n$, such that \begin{small} $\max (R = \mathbb{E}_{r_{i\ge 1}, x_{i\ge 1} \sim E_n, u_{i\ge 1} \sim \pi_{RL}}[R_1] )$\end{small} .
Unlike control algorithms, model-free reinforcement learning algorithms assume the system dynamic is unknown. Q-function, $Q^{\pi} (x_k, u_k)$ is defined as the expected return at state $x_k$  when we take action $u_k$ and adopt policy $\pi$   afterward:
 \begin{small} $Q^{\pi} (x_k, u_k) =  \mathbb{E}_{r_{i\ge k}, x_{i\ge k} \sim E_n, u_{i\ge k} \sim \pi}[R_k|x_k, u_k] )$\end{small}.
Q-learning   algorithms  \cite{watkins1992q} are among the most common   model-free RL methods for discrete action space problems.  These algorithms use the Bellman recursive equation
to model Q-function:
 \begin{small} 
$ Q^{\mu} (x_k, u_k) = 
\mathbb{E}_{r_{i\ge k}, x_{ i> k} \sim E_n}[r(x_k, u_k)+ \gamma Q^{\mu} (x_{k+1}, \mu(x_{k+1})) ] ),$
\end{small} 
where $\mu$ represents a greedy deterministic policy that selects the action which maximizes Q-value at each step:
 \begin{small} 
  \begin{equation}
\mu(x_k) = \text{argmax}_u Q(x_k,u_k). 
\label{argma}   
\end{equation}
 \end{small} 
Q-learning algorithms learn the parameters of the function approximator, $\theta^Q$,  by minimizing the Bellman error:
 \begin{small} 
$ \min(L(\theta^Q) = \mathbb{E}_{r_{k}, x_{ k} \sim E_n, u_k \sim \beta}[( Q (x_{k}, u_{k}|\theta^Q)-y_k)^2 ] )$, $
 y_k= r(x_k, u_k)+ \gamma Q(x_{k+1}, \mu(x_{k+1}))$,  
 \end{small} 
where $y_k$ is the fixed target Q-function, and $\beta$ represents the exploration policy.

For  continuous  action domain problems, it is not trivial to solve equation (\ref{argma})  at each time step. Finding an action to maximize $Q$ which can be a  complex nonlinear function  is computationally expensive or even infeasible. 
To address this problem, \cite{lillicrap2015continuous} proposed the Deep Deterministic Policy Gradient (DDPG) algorithm,  which   learns  two networks simultaneously. The critic network learns  Q-function by minimizing the Bellman error, and the actor network learns parameters of the policy  to maximize the estimated value of  Q-function. 
  Gu et al.\cite{gu2016continuous} proposed Normalized Advantage Function (NAF) Q-learning which formulates the Q-function as the sum of the value function, $V(x)$, and the advantage function, $A(x,u)$. 
   \begin{small} 
\begin{equation}
\begin{aligned}[c]
Q(x,u|\theta^Q)=V(x|\theta^V) + A(x,u|\theta^A),
\end{aligned}
\label{eq:AdvantagePlusValue}   
\end{equation}
 \end{small} 
where 
\begin{small} 
\begin{equation}
\begin{aligned}[c]
A(x,u|\theta^A)=-\frac{1}{2}(u - \mu(x|\theta^u))^TP(x|\theta^P)(u - \mu(x|\theta^u)).
\end{aligned}
\label{eq:Advantage}   
\end{equation}
 \end{small} 
\begin{small} $P(x|\theta^P)=L(x|\theta^P)L(x|\theta^P)^T$\end{small} , where \begin{small}  $L(x|\theta^P)$\end{small}    is a lower-triangular matrix. 
The value function is not a function of action, $u$. Therefore, the action which maximizes advantage function, $A$, maximizes the $Q$ function.  \begin{small} $P(x|\theta^P)$\end{small}   is a positive-definite matrix, and therefore, the action that maximizes the advantage function and the $Q$-function is given by \begin{small} $\mu(x|\theta^u)$\end{small} .




\section{Locally Linear Q-Learning}  
\label{sec:Locally Linear Q-Learning}

\begin{figure}[htbp]
\centerline{\includegraphics[scale=0.25]{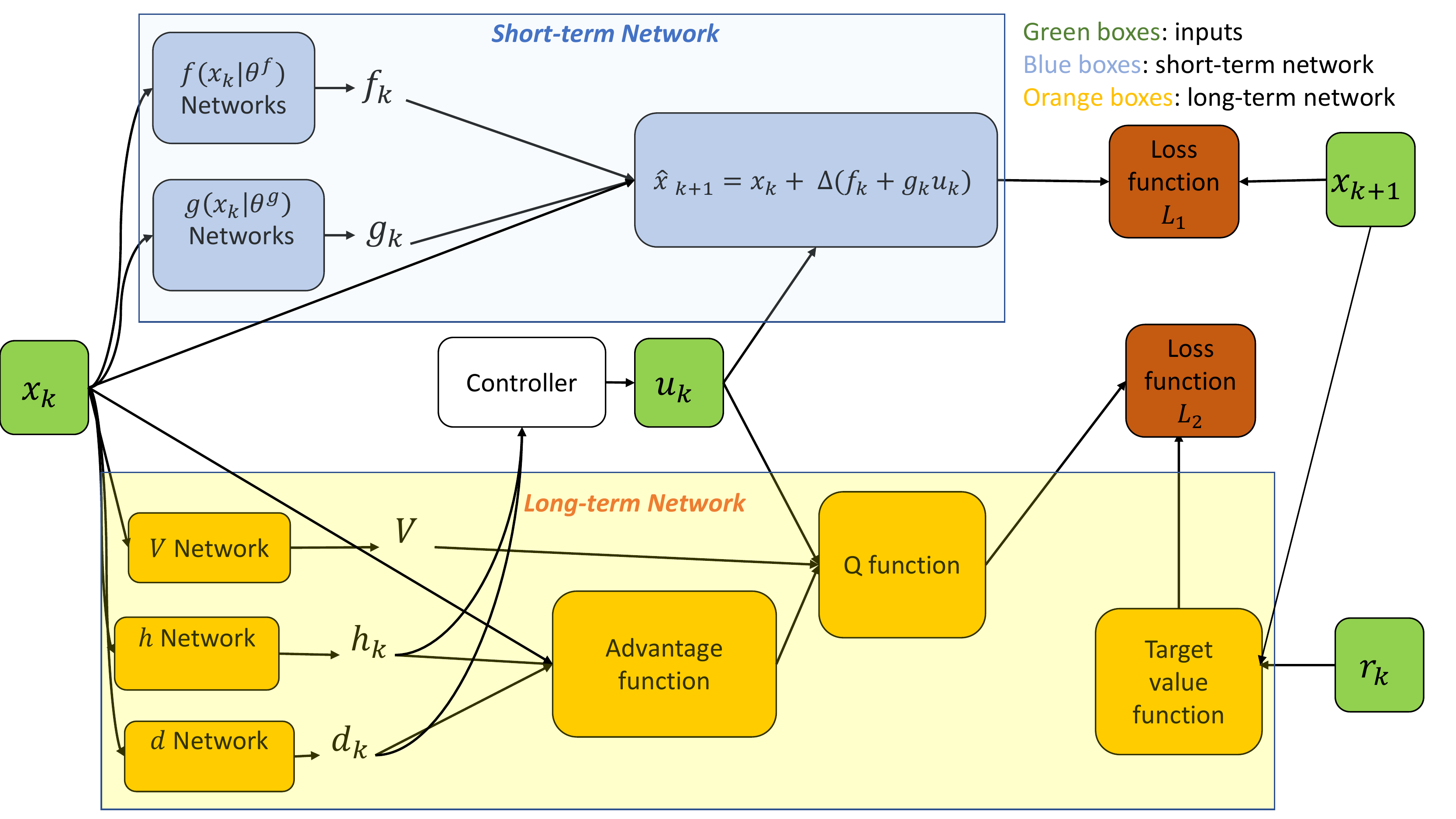}}
\caption{\small Learning the LLQL Network Parameters.   }
\label{LLQL}
\end{figure}
In this section, we propose  the LLQL algorithm, which like \cite{lillicrap2015continuous} and \cite{gu2016continuous} can handle continuous action space. Our approach learns short-term and long-term prediction models.  Using the long-term and short-term models, a controller  generates actions that guide the system toward its short-term and long-term goals. Figure \ref{LLQL} shows  our proposed structure to  learn the parameters of the short-term and  long-term prediction models. 

\textit{Short-term prediction:} consider the nonlinear  system  presented in equation (\ref{dis linear control}). 
In this work, we use deep neural networks to estimate system functions, $f(x_k)$, and $g(x_k)$ at each operating point. Substituting  the network estimations for these functions in equation (\ref{dis linear control}), we can predict the next state as: 
  \begin{small} 
\begin{equation}
\begin{aligned}[c]
\hat{x}_{k+1} =  x_k + \Delta (f(x_k|\theta^{f})  + g(x_k|\theta^{g})u_k),
\end{aligned}
\label{eq:dynamic1}   
\end{equation}
  \end{small} 
where   $\hat{x}_{k+1}$ represents our estimation of the next step, and $\theta^{f}$ and $\theta^{g}$ are the network parameters. $\Delta$ is a constant hyper parameter. In dynamic systems, the difference between two consecutive states, $x_{k+1} - x_k$, is typically very small. Considering a  small $\Delta$ leads to reasonable  $f$ and $g$ values and, therefore,     improves learning time and accuracy.

We call this dynamic system model \textit{short-term prediction model}. The controller uses this model to generate actions, which lead the system toward its short-term goals. Note that previous work  have used the system short-term dynamic model for generating additional samples in imagination rollout (for example, see \cite{gu2016continuous}, and \cite{racaniere2017imagination}). In this paper, we show that this model can also be used to design actions to achieve short-term goals. 
To learn the parameters of our short-term prediction model, $\theta^{f}$ and $\theta^{g}$, we minimize the short-term  loss function, $L_1$,  as it  is presented in Algorithm \ref{alg:LLQL}. 

\textit{Long-term prediction:}  Q-function represents the maximum cumulative  reward that can be achieved from current state, $x_k$, taking an action $u_k$. Therefore, by learning  Q-function,  we learn  the \textit{long-term prediction model} for the system. 
Like NAF~\cite{gu2016continuous}  (see equation (\ref{eq:AdvantagePlusValue})), we present Q-function as a sum of value function and advantage function. However, we present the     advantage function, $A(x,u|\theta^A)$ using a locally linear function of $x_k $ and $u_k$ as:
  \begin{small} 
\begin{equation}
\begin{aligned}[c]
Q(x,u|\theta^Q)=V(x|\theta^V) + A(x,u|\theta^A), \\
A(x,u|\theta^A)= -||(h(x_k|\theta^{h})+ d(x_k|\theta^{d}) u_k)||,
\end{aligned}
\label{eq:Qnewnew}   
\end{equation}
  \end{small} 
where $h(x_k|\theta^{h})$ and $d(x_k|\theta^{d})$ networks  model  the locally linear advantage function. 
Note the NAF advantage function is a special case of  the LLQL advantage  function when  $d(x_k|\theta^{d}) = I$, where $I$ represents the   identity matrix.

 To maximize Q-function, we have to  design $u_k$ which minimizes $h(x_k|\theta^{h})+ d(x_k|\theta^{d}) u_k$. For simplicity, we present  $h(x_k|\theta^{h})$, and $d(x_k|\theta^{d}) $ with $h_k$  and $d_k$ respectively in the remainder of the
  paper. To maximize  Q-function and achieve  the long-term goal, we can use  simple pseudo-inverse matrix multiplication and derive a solution with the least squares error as: 
    \begin{small} 
\begin{equation}
\begin{aligned}[c]
u_k=-(d_k^T d_k)^{-1} d_k^T h_k.
\end{aligned}
\label{eq:control}   
\end{equation}
  \end{small} 
When $||d_k||=0$, it means the network predicts that our action has no impact on the advantage function. Therefore, we choose a random action. Random exploration is an important part of any deep  RL algorithm. Therefore, in addition to this
 unlikely case, we add  noise, $\mathcal{N}_k$, to   the action, $u_k$,  during the training. We reduce the amount of noise injected to the action  as the algorithm converges.

\begin{footnotesize}
\begin{algorithm}
\small
	\def \MMM {\ensuremath{\mathcal{M}}\xspace}
	\def \DDD {\ensuremath{\mathcal{D}}\xspace}
	\def \UUU {\ensuremath{\mathcal{U}}\xspace}
	\newcommand{\Set}[1]{\ensuremath{\{#1\}}}
	\caption{Locally Linear Q-Learning Training}
	\label{alg:LLQL}
	\begin{algorithmic}[1]
		\STATE {Initialize Q network (equation (\ref{eq:Qnewnew})) with random weights.}
		\STATE {Initialize target network, $Q^{'}$, parameters: $\theta^{Q^{'}}= \theta^Q$. }
		\STATE {Create the reply buffer  $R = \emptyset$.}	
		\FOR{episode = 1:M}
		\STATE {Initialize a random process $\mathcal{N}$ for action exploration.  }
		\STATE {Receive  the initial observation, $x_0$.  }
		\FOR{k = 1:T}
		\IF{$||d_k||\ne0$}
			\STATE {Set $u_k =   -(d_k^T d_k)^{-1}(d_k^T ) h_k + \mathcal{N}_k$ }
		\ELSE
			\STATE{Set $u_k =   \mathcal{N}_k$ }
		\ENDIF
		\STATE {Execute $u_k$ and observe  $x_{k+1}$ and  $r_k$.}
		\STATE {Store transition $(x_k, u_k, x_{k+1}, r_k)$ in $R$.}
		\FOR{iteration = 1:$I_s$}
		\STATE {Randomly select a mini-batch of $N_s$ transition from $R$.}
		\STATE {Update $\theta ^ {f}$ and $\theta ^ {g}$ by minimizing the loss: $L_{1} =\frac{1}{N_s}\sum_{i=1}^{N_s}||x_{i+1}- x_i- \Delta( f(x_i|\theta^{f}) + g(x_i|\theta^{g}) u_i)||$.}
		\ENDFOR
		\FOR{iteration = 1:$I_l$}
		\STATE {Randomly select a mini-batch of $N_l$ transition from $R$.}
		\STATE {Set $y_i = r_i + \gamma Q^{'}(x_{i+1} |\theta^{Q^{'}})$.}
		\STATE {Update $\theta ^ Q$  by minimizing the loss: $L_2= \frac{1}{N_l}\sum_{i=1}^{N_l}||y_i-Q(x_{i}, u_{i}|\theta^Q)||$.}
		\STATE {Update the target  network: $\theta^{Q^{'}} = \tau\theta^{Q}  + (1-\tau)\theta^{Q^{'}} $}
		\ENDFOR
		\ENDFOR
		\ENDFOR
	\end{algorithmic}
\end{algorithm}    
\end{footnotesize}

In the application,  the controller solves  $u_k$ with  additional   constraints to achieve  the desired  short-term trajectories. We will discuss our  short-term adjustment algorithms in  the next section. 
To learn  Q-function, in addition to the  state estimation error, we  minimize the long-term loss function, $L_2$, as it  is presented in Algorithm \ref{alg:LLQL}. 
 Note that having the short-term model, it is straightforward to add imagination rollout to our algorithm to increase sample efficiency. However, improving sample efficiency in RL is not the focus of this work. 

\section{Control Strategy}
\label{sec:Short-term and long-term goals}

By separating  action design from prediction models, LLQL gives us the freedom to design different control strategies for achieving short-term and long-term goals. Moreover, the linear structure of short-term and long-term models simplifies the control design. 
Consider the case where   LLQL has learned a perfect long-term model for an environment using Algorithm \ref{alg:LLQL}. In this case, the optimum solution to achieve the long-term goal is given by equation (\ref{eq:control}). When we have one or more short-term goals as well,  we can formulate  the control design as an optimization problem to satisfy both short-term and long-term goals as much as possible.

In this paper, we consider two types of short-term goals: 1) desired trajectory, and 2) constraint. In the first scenario, the agent has a short-term desired trajectory. For example, a car may be required to travel with specific speed during certain periods. In the second scenario, the agent has some limitation for a specific period of time. For example, a car is required to  keep its speed below certain thresholds at some periods during the trip. To address the first problem, we add an additional term to the cost function for the short-term goal and solve for the action. We deal with  the second problem as a constraint optimization. 
\subsection{Short-term trajectory} 
Let $x_d$ represent our desired short-term trajectory. We  develop a  control strategy to track $x_d$ while pursuing the long-term goals. 
Using system dynamic functions  $f_k$ and $g_k$,  
we can define our  control optimization problem as: 
\begin{equation}
\begin{aligned}[c]
\begin{split}
\min_ \text{find $u_k$}(\gamma_1 (h_k+ d_k u_k)^2 + \gamma_2 (x^{d}_{k+1}-x_k- \Delta (f_k+g_k u_k))^2), \\
\end{split}
\end{aligned}
\label{eq: strategy}   
\end{equation}
 where $x^{d}_{k+1}$ represents the desired  trajectory at time $k+1$. $\gamma_1$ and $\gamma_2$ are positive coefficients and can be adjusted  to give higher weights to the short-term or long-term goals.  
 Note that in this work, we assume the short-term goals are temporary and when their time expires the system goes to the long-term optimum policy given by (\ref{eq:control}). 
 For example, we may require a car to have a specific speed  at some specific locations. 
 
We can apply a similar  pseudo-inverse matrix multiplication, and derive a solution with the least squares error for (\ref{eq: strategy}) as:
 \begin{small} 
 \begin{equation}
\begin{aligned}[c]
\begin{split}
u_k^*= ( \begin{bmatrix} 
\gamma_1   d_k \\
 - \gamma_2  \Delta g_k
\end{bmatrix} ^T
 \begin{bmatrix} 
\gamma_1   d_k \\
 - \gamma_2 \Delta g_k
\end{bmatrix}  )^ {-1} 
\end{split} \times \\
\begin{split}
 \begin{bmatrix} 
\gamma_1   d_k \\
 - \gamma_2 \Delta g_k
\end{bmatrix} ^T 
\begin{bmatrix} 
-\gamma_1   h_k \\
  \gamma_2 ( x_k+\Delta f_k -x^{d}_{k+1})
 \end{bmatrix}.
\end{split}
\end{aligned}
\label{eq: strategy solution}   
\end{equation}
\end{small}

 

 \subsection{Short-term constraint} 
 
 The LLQL algorithm provides a framework to design the actions considering different constraints. 
For safe operation,   the agent may have to  avoid  specific states for a period of time (for example, high speed or locations close to an obstacle). 
For simplicity, we assume at each moment we only have maximum one constraint on one state variable, $x^i$.  
This is a reasonable  assumption, because in physical systems the agent is close to one of the boundaries at any moment in time. When this is not the case, we can define new constraints as a combination of constraints. 
Consider $c^i_k$ as the constraint on the state variable, $x^i$,  at time $k$. 
We can define the constraint  optimization problem for LLQL as: 
\begin{equation}
\begin{aligned}[c]
\begin{split}
& \min_{\text{find} u_k} \frac{1}{2} (h_k+ d_k u_k)^2 \\ 
& \text{ such that: }  \\ 
& x^i_{k+1} \le c^i_{k+1}. \\
\end{split}
\end{aligned}
\label{eq:safefirst}   
\end{equation}
 $ \frac{1}{2}$ is a  coefficient  added to simplify the mathematical operation.  Using  our estimation of the next step, $x^i_{k+1} = x^i_k +\Delta ( f^i_k  + g^i_k u_k)$, we can derive the optimum action which satisfies the constraint as: 
 \begin{equation}
\begin{aligned}[c]
\begin{split}
&   u_{k}^*  =
-(d_k^T d_k)^{-1} d_k^T (h_k +  \lambda^* \alpha_1),
 \end{split}
\end{aligned}
\label{eq:constraintoptim}   
\end{equation}
where $\alpha_1 =\Delta {g^i_k}d_k^T (d_k d_k^T)^{-1}$,  $\alpha_2 =\Delta g^i_k(d_k^T d_k)^{-1} d_k^T $, and $ \lambda^*   = \frac{x^i_k + \Delta f^i_k - c_{k+1}  -\alpha_2h_k}{\alpha_1     \alpha_2}$. The derivation details for short-term 
constraints are presented in Section \ref{Derivation}.

\section{Application of LLQL Approximation to the State of the Art Deep RL Algorithms}
\label{Approximation}

Deep RL has made significant  progress in recent years. Algorithms such as Soft Actor-Critic (SAC) \cite{haarnoja2018soft}, Proximal Policy Optimization (PPO) \cite{schulman2017proximal}, Asynchronous Advantage Actor Critic (A3C) and Advantage Actor Critic (A2C) \cite{mnih2016asynchronous}, have made RL solutions more   general, more stable and   more sample efficient. In this section, we develop an approximation of LLQL algorithm which can be applied to any pre-trained RL solution. Using this  approximation, the users can apply the state-of-the-art RL algorithms + short-term adjustments. 

Consider equation (\ref{eq: strategy}). The first term in equation (\ref{eq: strategy}), $\gamma_1 (h_k+ d_k u_k)^2$, is designed to achieve the long-term goal. When we have a pre-trained RL model, the action generated by the network is designed to  achieve the long-term goal. Therefore, we can rewrite equation (\ref{eq: strategy}) by replacing  the first term in equation (\ref{eq: strategy}) by $\gamma_1 (u_k - u^{N}_k)^2$, where $u^{N}_k$ is the action generated by the pre-trained network.
The goal in this optimization problem is to minimize  the short-term error and the error between the new action and the action generated by the pre-trained network. Solving this optimization problem, we have:
 \begin{small} 
 \begin{equation}
\begin{aligned}[c]
\begin{split}
u_k^*= ( \begin{bmatrix} 
\gamma_1    \\
 - \gamma_2  \Delta g_k
\end{bmatrix} ^T
 \begin{bmatrix} 
\gamma_1    \\
 - \gamma_2 \Delta g_k
\end{bmatrix}  )^ {-1} 
\end{split} \times \\
\begin{split}
 \begin{bmatrix} 
\gamma_1    \\
 - \gamma_2 \Delta g_k
\end{bmatrix} ^T 
\begin{bmatrix} 
\gamma_1   u^{N}_k \\
  \gamma_2 ( x_k+\Delta f_k -x^{d}_{k+1})
 \end{bmatrix}.
\end{split}
\end{aligned}
\label{eq: strategy solution_1}   
\end{equation}
\end{small}
Note to compute  $u_k^*$ using equation (\ref{eq: strategy solution_1}), we do not need the long-term network. 
Similarly, we can replace $(h_k+ d_k u_k)^2$  in equation (\ref{eq:safefirst}) by $( u_k - u^{N}_k )^2$.
The goal of this new optimization problem is to minimize   the error between the new action and the action generated by the pre-trained network while satisfying the  constraints.
The optimum action which satisfies the constraint  can be computed as: 
 \begin{equation}
\begin{aligned}[c]
\begin{split}
&   u_{k}^*  =
u^{N}_k  -  \lambda^* \Delta {g^i_k},
 \end{split}
\end{aligned}
\label{eq:constraintoptim_2}   
\end{equation}
where  $ \lambda^*   = \frac{x^i_k + \Delta f^i_k - c_{k+1}  + \Delta {g^i_k} u^{N}_k }{\Delta {g^i_k}{\Delta {g^i_k}}^T}$. 

\section{Experimental Results}
\label{sec:Experimental Results}

\begin{figure}[t!]
     \centering
     \begin{subfigure}[b]{0.4\textwidth}
         \centering
         \includegraphics[width=\textwidth]{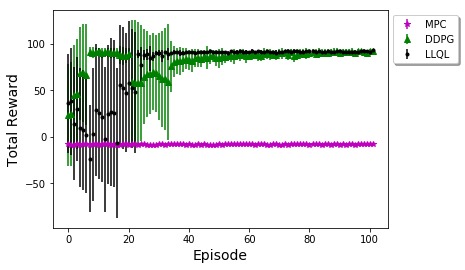}
         \caption{ \footnotesize {Average  and standard deviation of  cumulative rewards during the learning process. 
The experience stops after $1,000$ steps or when the car reaches the goal on top of the mountain, car's position $ = 0.5$, whichever occurs first. 
For each approach we performed the training 20 times, and selected  the top 5 models with the maximum cumulative rewards to calculate mean and standard deviation  of each episode. 
 } }
         
         \label{LearningjustRL}
     \end{subfigure}
     \hfill
     \begin{subfigure}[b]{0.4\textwidth}
         \centering
         \includegraphics[width=\textwidth]{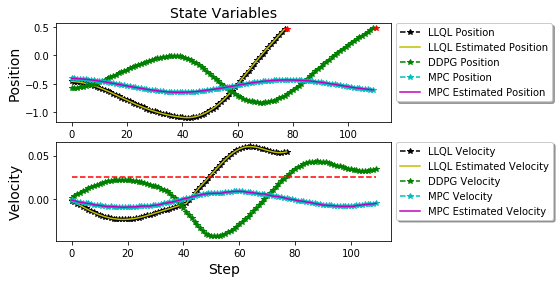}
         \caption{  \footnotesize { State variables and estimated state variables.
          Using LLQL model, the car reached  the top in 78 steps. Using DDPG model, the car reached the top in  110 steps. The car never reached the top when we used the MPC model. 
          The mean absolute error for  position, x, and  velocity, $v$, estimation in  each experiment are: $\bar{|e|}_{x_{LLQL}} = 0.00078$, $\bar{|e|}_{v_{LLQL}} = 0.000084$, $\bar{|e|}_{x_{MPC}} = 0.00039$, $\bar{|e|}_{v_{MPC}} = 0.00014$. For LLQL algorithm with short-term trajectory we set $\gamma_1=1$, $\gamma_2=2000$.
}}
         \label{justRL}
     \end{subfigure}
     \caption{\small {LLQL for MountainCarContinuous. The network's parameters are presented in Section \ref{NetworkParameters}.}}
        \label{LLQLMountainCarContinuous}
\end{figure}

In this section, we demonstrate the performance  of LLQL using  Mountain Car with Continuous Action (MountainCarContinuous) from OpenAI Gym\footnote{\url{http://gym.openai.com/envs/MountainCarContinuous-v0/}}. 
Figure \ref{LearningjustRL} shows the cumulative rewards during the training for the LLQL, the DDPG, and a model-based reinforcement learning based on MPC presented by  \cite{nagabandi2018neural}. 
The MPC based solution uses  the learned short-term predictive model (system dynamic model) to   generate  a sequence of actions that   maximize the reward over a finite horizon.  Figure \ref{justRL}  shows that the short-term predictive model estimates future states with high precision. However, 
 optimizing for a finite horizon is a  disadvantage for the model-based solution  in  achieving   long-term goals. Increasing the horizon may improve the long-term performance, but it also increases the computational costs in the application phase.  In our experiments, the  car never reached the top of the mountain using the model-based method.  Figure \ref{justRL}  presents the first 110 steps of a sample  experiment.   Unlike the MPC based solution, 
 the LLQL and the DDPG algorithms converged in less than 40 episodes (see Figure \ref{LearningjustRL}) and reach the top of the mountain in all experiments (see Figure \ref{justRL}).  


\subsection{Short-term trajectory}  

\begin{figure}[tp!]
     \centering
     \begin{subfigure}[b]{0.4\textwidth}
         \centering
         \includegraphics[width=\textwidth]{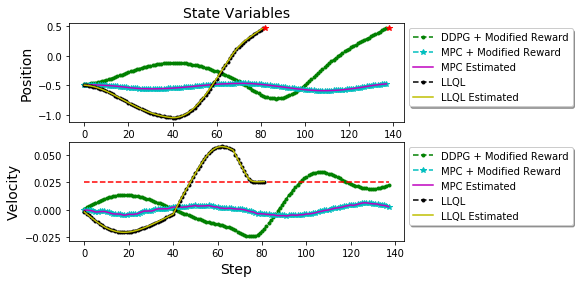}
         \caption{\footnotesize {With LLQL the car reached  the top of the mountain in 82 steps with a short-term goal error of $0.00008$. With DDPG + modified reward ($3^{rd}$ reward in Table \ref{Short}) the car reached  the top in 138 steps with short-term goal error $0.0023$. The MPC + modified reward  cannot guide the car to the top.}}
          \label{controlRL}
     \end{subfigure}
     \hfill
     \begin{subfigure}[b]{0.4\textwidth}
         \centering
         \includegraphics[width=\textwidth]{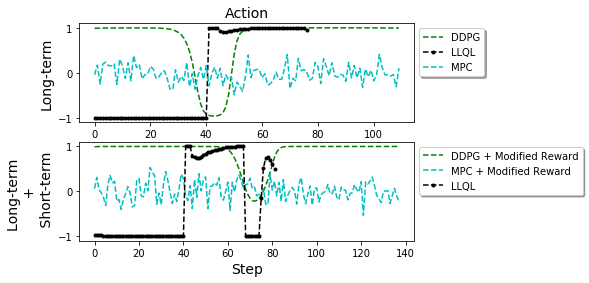}
         \caption{\footnotesize {Comparing  the actions in normal case versus  with the short-term trajectory. The hybrid strategy (see equation (\ref{eq:hybrid})) leads  to quick adjustment in LLQL action and therefore, smaller short-term goal error compared to the DDPG + reward modification solution.}}
         \label{Actiontr}
     \end{subfigure}
     \label{Shorterm}
     \caption{\small MountainCarContinuous with short-term and long-term goals. }
\end{figure}

Figure \ref{justRL} shows that the policy  presented in equation (\ref{eq:control})  can lead the car to the top of the mountain  using  the LLQL algorithm. 
We can see that the car's velocity is above $0.025$ (the red line) when it reaches  the top. Now consider the case where we want the car to reach  the top of the mountain with our desired velocity, $v_d = 0.025$. 
Using  equation (\ref{eq: strategy solution}), we can design a control strategy to reach   this goal without requiring  retraining the LLQL model. We apply the following hybrid control strategy to reach the top of the
 mountain with our desired speed. 
 \begin{small}
\begin{equation}
\begin{aligned}[c]
\begin{split}
& u_k = \begin{cases}
\text{use equation } (\ref{eq:control})& \text{if  $x_k < 0$} \\
\text{use equation } (\ref{eq: strategy solution}), & \text{otherwise.}
\end{cases}
 \end{split}
\end{aligned}
\label{eq:hybrid}   
\end{equation}
\end{small}
Figure \ref{controlRL} shows that the car can reach the top of the mountain with our desired velocity. When we did not impose   our desired speed to the system, the car reached the top of the mountain in 78 steps (see Figure \ref{justRL}). Demanding a lower speed, slowed down the car and  increased the number of steps to 82 (see Figure \ref{controlRL}).
Figure \ref{Actiontr} shows the actions with and without the short-term trajectory. 
We can see that the action temporarily becomes negative  to reduce the velocity to the desired level and then goes back to positive to push the car to the top of the mountain. 

 \begin{table}[htbp]
\footnotesize
\caption{\small Modified Reward functions for short-term trajectory.}
\begin{center}
\label{Short}
\begin{tabular}{||c||}
\hline
$r_{t1}= r_k -5000*|v_k-v_d|$ \text{if done}   \\
\hline
 $r_{t2}= r_k -100*|v_k-v_d|$ \text{if}   $x_k > 0.45$  \\
\hline
 \makecell{$r_{t3}= r_k -100*|v_k-v_d|$ \text{if}   $x_k > 0.45$ \\$r_{t3}= r_k -5000*|v_k-v_d|$ \text{if done} }   \\
\hline
$r_{t4}= r_k -25000*(v_k-v_d)^2$ \text{if done}\\
\hline

\hline
\end{tabular}
\end{center}
\end{table}
To solve this problem in the traditional way, we had to modify the reward function to achieve both short-term and long-term goals. For comparison,  we perform the following experience. We apply DDPG networks and MPC based networks  
with the modified reward functions shown in Table \ref{Short}. 
 Table \ref{Table_1} shows that the MPC based solution cannot guide the car to the top (success = 0/10). The DDPG 
with all 
 the modified reward 
  functions can achieve  the long-term goal in all the experiences, reaching the top of the mountain in 10 out of 10 experiments. However,  the DDPG based solutions  
  do not  perform very well with regard to the short-term goal (average absolute velocity error, $\bar{|e|_v}$, is too large). \textit{On the other hand, LLQL does not require   additional training or reward modification,   achieves  the long-term goal with average steps, $\bar{\text{s}}=89.5$,   and has the least velocity error, $0.4\%$.} 
\begin{table}[htbp]
\footnotesize
\caption{\small Short-term trajectory performance }
\scalebox{0.9}{
\begin{tabular}{ p{1.9cm}||p{1.8cm}||p{.6cm}||p{.5cm}||p{1cm} || }
 \hline
 \multicolumn{4}{|c|}{ LLQL Approximation Applied to   MountainCarContinuous } \\
 \hline
 \hline
 RL Algorithm &modified $r$  & $\bar{|e|_v} $ & $\bar{\text{s}}$ &  success\\
 \hline
\textit{DDPG}   &$r_{t1}$      &0.0232  &109.1&  10/10\\
\textit{DDPG} &  $r_{t2}$  &  0.0193 & 183.6 &10/10 \\
\textit{DDPG} &$r_{t3}$ &  0.0088 & 173.8& 10/10 \\
\textit{DDPG}   & $r_{t4}$   & 0.0193 &103.3 & 10/10\\
\textit{MPC} &   $r_{t1},r_{t2},r_{t3},r_{t4}$  &  - &1000 & 0/10 \\
\textbf{LLQL} & $-$ & \textbf{0.0001} & \textbf{89.5}  & \textbf{10/10} \\
 \hline
\end{tabular}}
 \label{Table_1}
\end{table}



\subsection{Short-term constraint}
Now consider the case where it is unsafe to drive the car above  a specific speed, for example, we plan to keep the speed under  $v_k \le 0.035$. We can use the following hybrid control strategy to achieve the long-term  goal while keeping the speed safe: 
\begin{small}
\begin{equation}
\begin{aligned}[c]
\begin{split}
& u_k = \begin{cases}
\text{use equation } (\ref{eq:control})& \text{if  $|v_k|$ $\le 0.033$} \\
\text{use equation } (\ref{eq:constraintoptim}), & \text{otherwise.}
\end{cases}
 \end{split}
\end{aligned}
\label{eq:hybrid2}   
\end{equation}
\end{small}
We selected the boundary slightly less than the hazardous threshold (0.033 instead of 0.035) to be safe. 
\begin{figure}[tp!]
     \centering
     \begin{subfigure}[b]{0.4\textwidth}
         \centering
         \includegraphics[width=\textwidth]{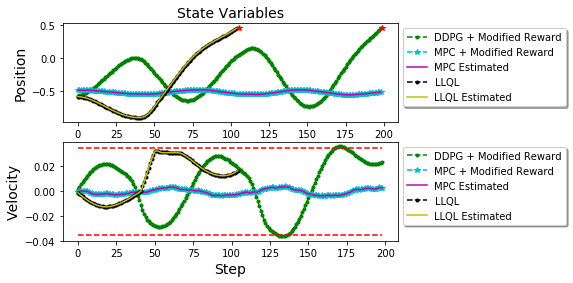}
         \caption{  \footnotesize {With LLQL the car reached  the top in 105 steps without violating the constraints. With DDPG + modified reward ($3^{rd}$ reward in Table \ref{tab2}) the car reached  the top in 199 steps and violated the constraints 13 timesteps. The MPC + modified reward  cannot guide the car to the top.} }
          \label{controlRL2}
     \end{subfigure}
     \hfill
     \begin{subfigure}[b]{0.4\textwidth}
         \centering
         \includegraphics[width=\textwidth]{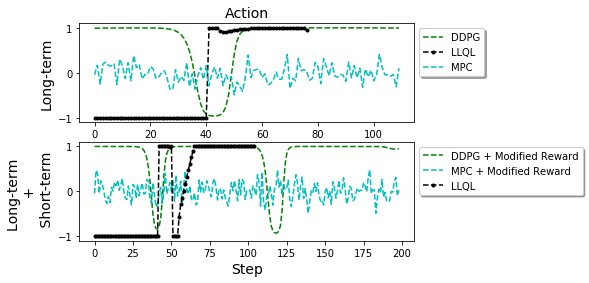}
         \caption{    \footnotesize {Comparing  the actions in the normal case versus  with the short-term constraint. The hybrid strategy (see equation (\ref{eq:hybrid2})) leads  to sharp  adjustment in LLQL action when the car gets close to the  hazardous areas. }}
         \label{Actionc}
     \end{subfigure}
     \label{Shorterm}
     \caption{\small MountainCarContinuous with long-term goal and short-term constraint. The  horizontal red lines represent the constraints. } 
\end{figure}
Figure \ref{controlRL2} shows that with the LLQL policy the car reaches its goal while staying outside of hazardous areas. The MPC based solution keeps the car outside of hazardous areas but cannot deliver the long-term goal (reaching the top of the mountain). The DDPG + modified reward reaches the top but fails to deliver the short-term goal (keeping the car out of hazardous areas).

\begin{figure}[t!]
     \centering
     \begin{subfigure}[b]{0.23\textwidth}
         \centering
         \includegraphics[width=\textwidth]{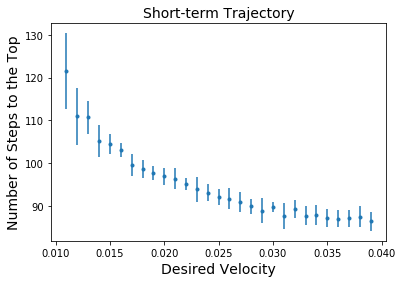}
         \caption{ \footnotesize {Number of steps to the top   vs desired  final velocity. }}
          \label{Trajeffect}
     \end{subfigure}
     \hfill
     \begin{subfigure}[b]{0.23\textwidth}
         \centering
         \includegraphics[width=\textwidth]{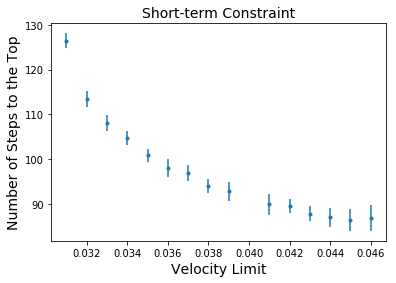}
         \caption{  \footnotesize{Number of steps to the top   vs  velocity constraint. }}
         \label{ConEffect}
     \end{subfigure}
     \label{Effect}
     \caption{\small Long-term performance vs short-term goals. We run the model with each short-term goal 10 times and present the average and standard deviation of the long-term goal.}
\end{figure}

 \begin{table}[htbp]
\footnotesize
\caption{\small Modified Reward functions for constraint performance.}
\begin{center}
\begin{tabular}{||c||}
\hline
$r_{c1}= r_k - 10$  \text{if}   $|v_k| > 0.033$   \\
\hline
  $r_{c2}=r_k - 100 (|v_k|-0.033)$  \text{if}   $|v_k| > 0.033$   \\
\hline
$r_{c3}= r_k - (100(|v_k|-0.033))^2$  \text{if}   $|v_k| > 0.033$    \\
\hline
$r_{c4}=-10$  \text{if}   $|v_k| > 0.033$ \\
\hline

\hline
\end{tabular}
\label{Short_2}
\end{center}
\end{table}
\begin{scriptsize}
\begin{table}[htbp]
\footnotesize
\caption{\small Short-term constraint performance}
\scalebox{0.9}{
\begin{tabular}{ ||p{1.9cm}||p{1.8cm}||p{.6cm}||p{.5cm}||p{1cm} || }
 \hline
 \multicolumn{4}{|c|}{ LLQL Approximation Applied to   MountainCarContinuous } \\
 \hline
 \hline
 RL Algorithm &modified $r$  & $\bar{s}_{out} $ & $\bar{\text{s}}$ &  success\\
 \hline
\textit{DDPG}   &$r_{c1}$      &21.5  &100&  10/10\\
\textit{DDPG} &  $r_{c2}$  &  25.2& 106.8 &10/10 \\
\textit{DDPG} &$r_{c3}$ &  20.4 & 147.5& 10/10 \\
\textit{DDPG}   & $r_{c4}$   & 25.1 &104.1& 10/10\\
\textit{MPC} &   $r_{c1},r_{c2},r_{c3},r_{c4}$  &  0 &1000 & 0/10 \\
\textbf{LLQL} & $-$ & \textbf{0} & \textbf{98.9}  & \textbf{10/10} \\
 \hline
\end{tabular}
}
 \label{tab2}
\end{table}
\end{scriptsize}
Like the previous section, we apply DDPG network + modified reward function and MPC  + modified reward function (see Table \ref{Short_2}) to compare LLQL with the traditional model-free and model-based reward engineering approaches.   
Table \ref{tab2} shows that unlike LLQL, the modified 
  rewards fail to keep the car below the allowed speed while reaching the top of the mountain.
  Using equations (\ref{eq: strategy solution}) or (\ref{eq:constraintoptim}) for deriving a set of actions  is equivalent to solving a sub-optimum solution for  the long-term goal in order to  satisfy the  short-term desired trajectories or constraints. When  the short-term goals are far from  the global optimum solution, the long-term performance degrades. Figure \ref{Trajeffect} shows  that lower desired velocities  lead to longer traveling time for the MountainCar. Similarly,  Figure \ref{ConEffect} shows that further limiting the maximum velocity degrades the long-term performance. 

\subsection{Application of LLQL Approximation to  Pre-trained Deep RL Algorithms}
\label{ApplicationApproximation}
We use three pertained models from Stable Baselines (SB) \cite{stable-baselines} for the Pendulum-v0\footnote{\url{http://gym.openai.com/envs/Pendulum-v0/}} to show that LLQL's approximation can be used to achieve new adjustments  in pre-trained models (see details in Table \ref{Pendulum_1}, Table \ref{Pendulum_2}). 
For the short-term trajectory control, we set the desired velocity at $v_d=0$ when the Inverse Pendulum is vertical, $cos(x)>0.99$. Note that in this case the short-term trajectory is in harmony with the long-term goal and, therefore, as we can see in Table \ref{Pendulum_1} even though the short term desired trajectory is improved significantly, the average absolute error, $\bar{|e|_v}$, is significantly lower when we apply  the adjustment, the average  cumulative reward, $\bar{R}$ is not affected in a negative way.  
For the short-term constraint control, we set the maximum velocity at $v=6$. This is significantly lower  than the Inverse Pendulum maximum velocity, $v_{max} =8$. Table \ref{Pendulum_2} shows that the approximation of the LLQL can be used to satisfy new constraints:  number of steps out of speed limit is zero when we apply the adjustment, $ \bar{s}_{out} =0$. In this experiment, we used the same $f$  and $g$ network architectures and learning meta parameters  that we used for the MountainCarContinuous.  

\begin{table}[htbp]
\footnotesize
\caption{\small Short-term trajectory: $v=0$ if $cos(x)>.99$ }
\scalebox{0.9}{
\begin{tabular}{ ||p{1.1cm}||p{.8cm}||p{1.8cm}||p{1cm}||p{1.6cm}  ||}
 \hline
 \multicolumn{4}{|c|}{ LLQL Approximation Applied to   Pendulum-v0  } \\
 \hline
 \hline
 Algorithm & \small \makecell{ $\bar{|e|_v}$}   & $\bar{|e|_v} \text{+ adjustment}$ & $\bar{R}$ &  $\bar{R} \text{+  adjustment}$\\
 \hline
SAC \cite{haarnoja2018soft}   & 0.03    &0.003  &-162 &  -151\\
PPO \cite{schulman2017proximal} &   0.16  &  0.008 &-173 & -172 \\
A2C  \cite{mnih2016asynchronous} & 0.03 & 0.005 & -164& -173 \\
 \hline
\end{tabular}
\label{Pendulum_1}
}
\end{table}

\begin{table}[htbp]
\footnotesize
\caption{\small Short-term constraint: $v<6$ }
\scalebox{0.9}{
\begin{tabular}{|| p{1.1cm}||p{.7cm}||p{2.1cm}||p{.7cm}||p{1.9cm} || }
 \hline
 \multicolumn{4}{|c|}{ LLQL Approximation Applied to   Pendulum-v0  } \\
 \hline
 \hline
 Algorithm &$\bar{s}_{out}$  & $\bar{s}_{out}  \text{+  adjustment}$& $\bar{R}$ &  ${\bar{R}  \text{+  adjustment}}$\\
 \hline
  SAC \cite{haarnoja2018soft}  & 0.40  & 0 &-162 &   -170\\
PPO \cite{schulman2017proximal} & 0.31 &  0  & -173  &-184\\
A2C  \cite{mnih2016asynchronous}  &0.21 & 0 & -164&  -167\\
 \hline
\end{tabular}
\label{Pendulum_2}
 }
\end{table}

 \section{Conclusions}
 \label{sec:Conclusions}
  In this work, we presented LLQL as a new model-based RL algorithm with  the capability of achieving  both short-term and long-term goals without requiring complex reward functions. 
Moreover, we presented an approximation of LLQL which can be applied to any pre-trained RL algorithm. 
This can be very significant for industrial applications where the RL algorithms have  not been used due to the necessity of  different short-term adjustments. In the future work, we will investigate conditions where short-term goals are feasible and develop a more analytical approach to set the meta parameters for the controller to guarantee short-term and long-term goals. Moreover, we will model uncertainties in short-term prediction model and apply robust control theory to design robust control solutions.
\bibliographystyle{IEEEtran}

\bibliography{references}

\appendices

\section{Derivation Details for Short-term Constraints}
\label{Derivation}

Consider equation (\ref{eq:safefirst}). Substituting  our estimation of the next step from equation  (\ref{eq:dynamic1})  in (\ref{eq:safefirst}), we have 
 
 \begin{equation}
\begin{aligned}[c]
\begin{split}
& \min_{\text{find} u_k} \frac{1}{2} (h_k+ d_k u_k)^2 \\ 
& \text{ such that: }  \\ 
& x^i_k + \Delta(f^i_k  + g^i_k u_k) \le c^i_{k+1}  \\
\end{split}
\end{aligned}
\label{eq:safe1}   
\end{equation}
 
Using Lagrangian method at each time step $k$, we have
\begin{equation}
\begin{aligned}[c]
\begin{split}
& L(u_{k}, \lambda)= \frac{1}{2}  (h_k+ d_k u_k)^2 +  \\  
& \lambda (x^i_k + \Delta(f^i_k  + g^i_k u_k )- c^i_{k+1}  ) \\
\end{split}
\end{aligned}
\label{eq:lagra}   
\end{equation}

Taking the gradient of $L$ with respect to $u_{k}$, we can write  the Karush-Kuhn-Tucker (KKT) \cite{kuhn2014nonlinear} conditions for optimal solution of equation (\ref{eq:safe1}), $\{u_{k}^*, \lambda^*\}$ as: 
\begin{equation}
\begin{aligned}[c]
\begin{split}
& (h_k+ d_k u_k^*) d_k + 
  \lambda^* {\Delta g^i_k}= 0\\ 
 & \lambda^* (x^i_k + \Delta( f^i_k  + g^i_k u_k^* ) -  c_{k+1} ) = 0 
\end{split}
\end{aligned}
\label{eq:safe}   
\end{equation}

With this assumption, we can show 
\begin{equation}
\begin{aligned}[c]
\begin{split}
&   u_{k}^*  =
-(d_k^T d_k)^{-1} d_k^T (h_k +  \lambda^* \Delta {g^i_k}d_k^T (d_k d_k^T)^{-1})
 \end{split}
\end{aligned}
\label{eq:constraint3}   
\end{equation}

Note that when there is no constraint: $\lambda^* =0$, 
we have $ u_{k}^* = -(d_k^T d_k)^{-1} d_k^T h_k$. This is exactly the input we computed in equation (\ref{eq:control}). 
When $\lambda^* \ne 0$, we have
\begin{equation}
\begin{aligned}[c]
\begin{split}
&  x^i_k +\Delta( f^i_k  + g^i_k u_k^* ) -  c_{k+1} =0
 \end{split}
\end{aligned}
\label{eq:con}   
\end{equation}

We define  $\alpha_1 =\Delta {g^i_k}d_k^T (d_k d_k^T)^{-1}$, and  $\alpha_2 =\Delta g^i_k(d_k^T d_k)^{-1} d_k^T $. $\alpha_1$ and  $\alpha_2$ are scalar.
Substituting $ u_{k}^*$ from equation (\ref{eq:constraint3}) in equation (\ref{eq:con}) we have 
\begin{equation}
\begin{aligned}[c]
\begin{split}
&  x^i_k + \Delta f^i_k  - \alpha_2  (h_k +  \alpha_1 \lambda^*) -  c_{k+1} =0
 \end{split}
\end{aligned}
\label{eq:sol}   
\end{equation}
Therefore,
\begin{equation}
\begin{aligned}[c]
\begin{split}
&   \lambda^*   = \frac{x^i_k + \Delta f^i_k - c_{k+1}  -\alpha_2h_k}{\alpha_1     \alpha_2} \\
& u_{k}^*  =
-(d_k^T d_k)^{-1} d_k^T (h_k +  \lambda^* \alpha_1).
 \end{split}
\end{aligned}
\label{eq:sol}   
\end{equation}

\section{Networks Parameters}
\label{NetworkParameters}
In all the networks we shift and scale the state variables to zero mean and unit standard deviation for a better learning. For exploration, we add an additive normal noise to the action. We reduce the noise after each episode with positive cumulative rewards.  All the activation functions are Rectified Linear Units (ReLUs).
We used the following network structures and parameters in for each model. 
\begin{itemize}
\item \textit{LLQL:} $h$, $d$, $V$, $f$  and $g$ networks each has two hidden layers with 200 neurons in each layer. 
 The number of iterations for short-term and long-term prediction model: $I_s = I_l=5$. The learning rate for the long-term prediction model  is  $0.001$. 
         The batch size for this model is 10.  The discount rate
  $\gamma = 0.999$. The target model update rate, $\tau = 0.001 $. The learning rate for the  short-term prediction model is $0.001$ for the first 20000 steps and then reduces to $0.0001$.   $\Delta = 0.001$. The batch size for this model  is 100.

\item \textit{MPC:}  we use $f$  and $g$ networks for state prediction. 
  
\item \textit{DDPG:}   The Q-network, and the 
 deterministic policy network each has  two hidden layers with 200 neurons.  
 The learning rate for the Q-network is  $0.00001$, and the learning rate for the deterministic policy network is $0.000001$. The discount rate
  $\gamma = 0.99$. The batch size is 8.  The  target model update rate for both networks is $ 0.1$. 
  \end{itemize} 

\end{document}